\documentclass[10pt,twocolumn,letterpaper]{article}

\usepackage{cvpr}
\usepackage{times}
\usepackage{epsfig}
\usepackage{graphicx}
\usepackage{amsmath}
\usepackage{amssymb}
\usepackage[]{algorithm2e}
\usepackage[pagebackref=true,breaklinks=true,letterpaper=true,colorlinks,bookmarks=false]{hyperref}

 \cvprfinalcopy % *** Uncomment this line for the final submission

\def\cvprPaperID{1528} % *** Enter the CVPR Paper ID here

\ifcvprfinal\fi
\begin{document}

%%%%%%%%% TITLE
\title{OCGAN: One-class Novelty Detection Using GANs with Constrained Latent Representations}

\author{Pramuditha Perera\thanks{This work was conducted as part of the first author's internship at AWS AI.}\\
Johns Hopkins University\\
{\tt\small pperera3@jhu.edu}
% For a paper whose authors are all at the same institution,
% omit the following lines up until the closing ``}''.
% Additional authors and addresses can be added with ``\and'',
% just like the second author.
% To save space, use either the email address or home page, not both
\and
Ramesh Nallapati\\
AWS AI\\
{\tt\small rnallapa@amazon.com}
\and
Bing Xiang\\
AWS AI\\
{\tt\small bxiang@amazon.com}
}

%\and
%Ramesh Nallapati\\
%Amazon AI\\
%{\tt\small rnallapa@amazon.com}

%\and
%Bing Xiang\\
%Amazon AI\\
%{\tt\small bxiang@amazon.com}
%}

\maketitle
%\thispagestyle{empty}

%%%%%%%%% ABSTRACT
\begin{abstract}
We present a novel model called {\it OCGAN} for the classical problem of one-class novelty detection, where, given a set of examples from a particular class, the goal is to determine if a query example is from the same class. Our solution is based on learning latent representations of in-class examples using a denoising auto-encoder network. The key contribution of our work is our proposal to explicitly constrain the latent space to {\it exclusively} represent the given class. In order to accomplish this goal, firstly, we force the latent space to have bounded support by introducing a {\it tanh} activation in the encoder's output layer. Secondly, using a discriminator in the latent space that is trained adversarially, we ensure that encoded representations of in-class examples resemble uniform random samples drawn from the same bounded space. Thirdly, using a second adversarial discriminator in the input space, we ensure all randomly drawn latent samples generate examples that look real. Finally, we introduce a gradient-descent based sampling technique that explores points in the latent space that generate potential out-of-class examples, which are fed back to the network to further train it to generate in-class examples from those points. The effectiveness of the proposed method is measured across four publicly available datasets using two one-class novelty detection protocols where we achieve state-of-the-art results. 

\end{abstract}

\section{Introduction}

One-class novelty detection tackles the problem of quantifying the probability that a test example belongs to the distribution defined by training examples \cite{dsvdd}. Different from other machine learning tasks, in one-class novelty detection, examples of only a single class are observed at training time. During inference, the trained model is expected to accept \textit{in-class} examples and reject \textit{out-of-class} examples. Since the problem formulation assumes unavailability of any negative training data, it is a difficult problem to solve in practice. Nevertheless, it has a number of applications including abnormality detection \cite{Saleh:2013:OAD:2514950.2516141},\cite{oza2019one} intruder detection \cite{ocfeatures},\cite{oza2019active},\cite{btasoc}, bio-medical data processing \cite{roberts} and imbalance learning \cite{Napierala2016}.

\begin{figure}
    \centering
\includegraphics[width=1\linewidth]{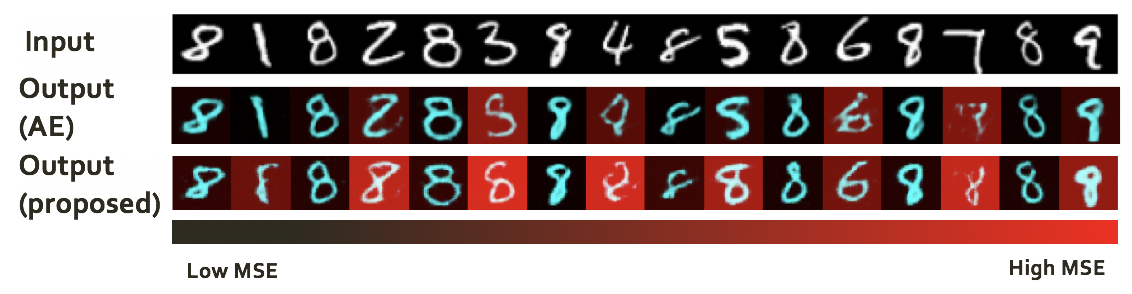}\hskip30pt
    \caption{Limitations of in-class representation based novelty detection. Top: Input images; Middle: Output of an auto-encoder network trained on digit 8. Bottom: Output produced by OCGAN, the proposed method. Even though auto-encoder network is trained only on digits of 8, it provides good reconstruction for digits from classes 1,5,6 and 9. In contrast, OCGAN forces the latent representation of any example to reconstruct a digit 8. As a result, all out-of-class examples produce high Mean Squared Error (MSE). The intensity of red color in the bottom two rows is proportional to the MSE.}
    \label{fig:comp}
\end{figure}

With the advent of deep learning, one-class novelty detection has received considerable amount of attention in the literature. Contemporary works in one-class novelty detection focus on learning a representative latent space for the given class  \cite{AE,GPND}. Once such a space is learned, novelty detection is performed based on the projection of a query image onto the learned latent space. Two distinct strategies are commonly used for this purpose in the literature. In the first strategy, the difference between the query image and its inverse image (reconstruction) is used as a novelty detector. Various distance measures ranging from mean squared error \cite{AE} to discriminator output \cite{cvpr2018} have been used in the literature for this purpose. In comparison, the second strategy explicitly models the learned latent space using a distribution \cite{AND,GPND,dsvdd,btasoc}. In this work, we consider the former strategy for novelty detection. We investigate limitations of existing representation learning techniques and propose learning a latent space that {\it exclusively} generates only in-class examples, to improve performance in novelty detection.

Existing work focuses on generating a latent representation that preserves details of the given class. In doing so, it is assumed that when an out-of-class object is presented to the network, it will do a poor job of describing the object, thereby reporting a relatively higher reconstruction error. However, this assumption does not hold at all times. For an example, experiments done on digits in the literature \cite{GPND,AND} suggest that networks such as auto-encoders trained on digits with a simple shape such as 0 and 1 have high novelty detection accuracy. In contrast, digits with complex shapes, such as digit 8, have relatively weaker novelty detection accuracy. This is because a latent space learned for a class with complex shapes inherently learns to represent some of out-of-class objects as well. As an example, the latent space learned on digit 8 is also able to represent other digits such as 1,3,6,7 reasonably well -- thereby producing very low distance error values for out-of-class examples as shown in Figure~\ref{fig:comp} (middle). 

We note that the requirement in novelty detection is not only to ensure that in-class samples are well represented; it is also to ensure that out-of-class samples are poorly represented. To the best of our knowledge, none of the previous work has addressed the latter requirement. In this work, we propose One-Class GAN (OCGAN), a two-fold latent space learning process that considers both these requirements. 

At a high-level, we learn a latent space that represents objects of a given class well. Secondly, we ensure that any example generated from the learned latent space is indeed from the known class. In other words, if the network is trained on a digits of 8, we ensure that any sample drawn from the latent space, when used to generate an image, corresponds to an image of digit 8. This ensures that out-of-class samples are not well represented by the network.  Shown in Figure~\ref{fig:comp}(bottom) are the outputs generated by the proposed method for the inputs shown in Figure~\ref{fig:comp}(top). Since the entire latent space corresponds to images from digit 8, all projections into the latent space in return produce images of digit 8. %This contribution makes our method unique from all previous novelty detection methods in the literature. 

%As evident from Figure~\ref{fig:comp}(bottom), network forces images of all classes to look 

 %To summarize, we make following contributions in this paper:
%\begin{itemize}
%    \item We bound the support of the latent space using a {\it tanh} function to facilitate thorough sampling from the latent space. Further, we force given class to evenly populate the latent space using a latent discriminator that distinguishes between latent projections of real examples and samples drawn uniformly at random from latent space.
%    \item We introduce an object-level discriminator to force objects corresponding to random latent space samples to look similar to objects of the given class.
%    \item A object-level classifier, that is different from learned discriminators, is used to assess the quality of generated images from latent-space samples. We propose to use feedback of this classifier to mine sub-spaces in the latent space that may result in out-of-class reconstruction. 
%\end{itemize}

\section{Related Work}
\noindent{\textbf{One-class Novelty Detection.}} One-class novelty detection is a well-defined research problem with standard evaluation metrics that has received considerable attention in the recent past. It has been traditionally treated as a representation-learning problem. The earliest methods in one-class novelty detection used Principal Component Analysis (PCA) \cite{Bishop:2006:PRM:1162264} and its kernel extension \cite{HOFFMANN2007863} to find a subspace that best describes the given concept. With the advent of neural networks and deep learning, a similar mapping was sought using auto-encoder networks \cite{MNISTAUTO}. 

As discussed in the preceding section, once such a mapping is learned, one-class novelty detection is carried out either based on reconstruction error or by explicitly modeling the normal behaviour of the known class in the latent space. In \cite{HOFFMANN2007863} and \cite{AE} the former strategy has been used to perform novelty detection using mean squared error as the novelty function. In \cite{cvpr2018}, a Generative Adversarial Network (GAN) \cite{goodfellow2014generative}  is trained to de-noise noisy samples of the given class. There, the discriminator's prediction in the image space is used to quantify reconstruction error. Following a slightly different strategy, \cite{IPMI} proposes to learn a mapping between a random distribution and the image manifold of the given class. In \cite{IPMI}, the closest image to a query is sought through back-propagation, where novelty detection is performed based on the difference between the two images.

The latter strategy, where the behavior of the known class in the latent space is modeled, has also received considerable attention in recent works. Earlier work of this nature used one-class modeling tools such as One-class SVM \cite{Scholkopf:2001:ESH:1119748.1119749} and Support Vector Data Descriptor (SVDD) \cite{Tax:2004:SVD:960091.960109} on top of an obtained latent representation. In \cite{GPND}, first, a GAN is used to obtain a latent representation. Then, the probability distribution of the latent space is modeled as a product of two marginal distributions where marginal distributions are learned empirically. In contrast, in \cite{AND} the latent distribution is modeled using an auto-regressive network that is learned along with the parameters of the auto-encoder. Using a different approach, deep-SVDD \cite{dsvdd} tries to learn a latent space where intra-class variance is low. The method proposed by \cite{dsvdd} is conceptually similar to \cite{ocfeatures} but does not use any external data in finding the solution as done in the latter work.

\noindent{\textbf{Anomaly Detection and One-class Classification.}} Both anomaly detection \cite{ICCVoutlier} and one-class classification \cite{Scholkopf:2001:ESH:1119748.1119749} are problems related to one-class novelty detection. Both have  similar  objectives -- to detect out-of-class samples given a set of in-class samples. A hard label is expected to be assigned to a given image in one-class classification; therefore its performance is measured using detection accuracy and F1 score. In contrast, novelty detection is only expected to associate a novelty score to a given image; therefore performance of novelty detection is measured using a Receiver Operating Characteristic (ROC) curve. However, boundary-based one-class classification methods such as One-class SVM, and SVDD can be adopted as novelty detection methods by considering distance to the decision boundary as the novelty score. In contrast, anomaly detection (also known as outlier detection) is an unsupervised learning task \cite{ICCVoutlier}. Given a mixture of unlabeled in-class and out-of-class examples, goal of anomaly detection is to separate in-class examples from out-of class examples. Since anomaly detection and novelty detection follow different protocols, we note that these two tasks are not comparable. Therefore, tools designed for anomaly detection and novelty detection cannot be used interchangeably.

\noindent{\textbf{Adversarial Learning.}} Given a set of images, Generative Adversarial Networks introduced in \cite{goodfellow2014generative} play a two-player game between a generator network and a discriminator network. Here, the generator network tries to produce realistic images (fake images) from the given image distribution whereas the discriminator network tries to distinguish fake images from real images. At equilibrium, the generator network learns the distribution of the given image set. In order to achieve this state, GAN theory dictates that there should be a balance between the capacities of the two networks. In \cite{mirza2014conditional}, GAN was extended to the conditional setting. Based on this extension, GANs have been used in many image-to-image translation applications since. It was shown in \cite{DCGAN} that GANs can be used to learn stable representations even with deep convolutional networks, provided that certain design choices are made. Inspired by the network architecture of \cite{cvpr2018}, and following principles outlined in \cite{DCGAN}, we propose a deep convolutional GAN architecture as the back-bone of our solution. 

\section{Proposed Method: OCGAN}

%In this section, first we present the motivation of the proposed solution. Then we describe each component of our solution with architectural details. 

\subsection{Motivation}

In the introduction, we presented an example where a network trained to represent a given class has ended up providing good representation for images of other classes. When images of a given class are sufficiently diverse, smoothly transitioning between the projection of one in-class image in the latent space to that of another can be done along infinitely many different paths -- this is particularly the case for latent spaces with high dimensionality. In training auto-encoders, we model projections of only observed examples into the latent space - not all possible paths between the corresponding latent points.

\begin{figure}
    \centering
\includegraphics[width=1\linewidth]{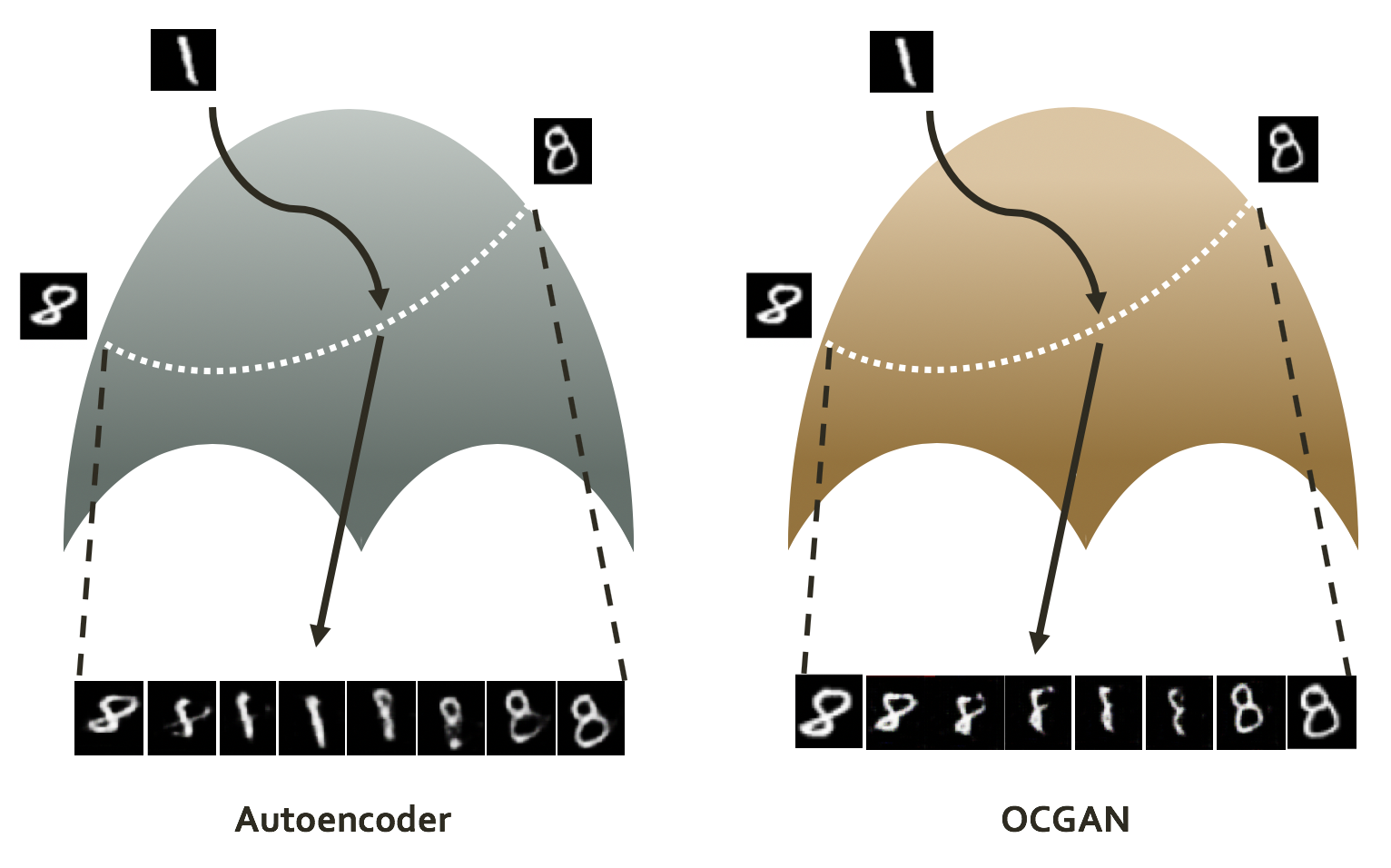}\hskip30pt
    \caption{ This figure illustrates the latent space learned for digit 8 using a denoising-autoencoder network (left) and the proposed method (right). Visualization of a chosen path between two digit images in the latent space are shown in the figure. In the denoising auto-encoder, digit 1 is well represented in this path. Therefore, for a digit 1 image, the reconstruction error will be low. On the other hand, the proposed method produces in-class examples throughout the chosen path in the latent space between the two images. Therefore, when a digit 1 image that gets projected into this path is considered, we find that reconstruction error is high.}
    \label{fig:latentcomp}
\end{figure}

In Figure~\ref{fig:latentcomp} we visualize a path traced in the latent space between two points corresponding to two different images of the given class (class 8). This visualization reveals that as we transition from one point to the other in the latent space along the specified path, certain intermediate latent samples resemble the likeness of digit 1. When the network observes an instance of digit 1, it gets projected onto such samples. Since digit 1 is well represented by the network, its reconstruction error will be low, although it is out of class. The core idea of our proposal is based on this observation. We argue that if the entire latent space is constrained to represent images of the given class, the representation of out-of-class samples will be minimal -- thereby producing high reconstruction errors for them.

With this strategy in mind, we explicitly force the entirety of the latent space to represent only the given class. When applied to the example in Fig. \ref{fig:latentcomp}, all latent samples along any path between the two 8's will reconstruct into a set of digit 8 images. Visualization of the path as shown in Figure~\ref{fig:latentcomp}(b) validates this claim. As a result, when an out-of-class digit 1 is presented to the model, there will be a high difference between the digit and the reconstruction of the digit (which will now look more like a digit 8). As a result, the proposed method is able to produce superior novelty detection performance.

\subsection{Proposed Strategy}
The proposed solution, OCGAN, consists of four components: a denoising auto-encoder, two discriminators (latent and visual discriminator) and a classifier. The proposed network is trained using adversarial principles. We describe each of these components in detail below.

%\begin{itemize}
 %\noindent 1. Learn a network with good in-class representation. We use a denoising auto-encoder to learn such a representation.

%\noindent 2. Force in-class examples to populate the latent space uniformly using a latent adversarial discriminator that tries to distinguish between latent projections of in-class examples and samples drawn from the latent space uniformly at random.

%\noindent 3. Ensure any latent sample corresponds to images of the known class. An adversarial visual discriminator, that tries to distinguish between real in-class images and images generated from random samples in the latent space, is used to ensure that any random latent-space sample ends up generating an image of the given class. 

%\noindent 4. Seek samples from the latent space that generate poor quality images from the given class to make the learning process stronger. We train a classifier to facilitate mining such samples, which we call {\it informative negatives}.
%\end{itemize}

 %By following this strategy, we ensure that not only the given class is represented well by the network, but also that entirety of the latent space is designated to represent the given class. We achieve this by randomly sampling from the latent space and forcing its decoded image to belong to the given class. When the dimension of latent space is too high, we use a hard-negative sample mining strategy to perform sampling. 
 %In following sub-sections we describe the proposed strategy in detail.
 
\noindent{\textbf{Denoising auto-encoder:}} Following previous work, we use a denoising auto-encoder network to learn a representation for the given concept. The auto-encoder is an encoder (En) - decoder (De) structure that is trained with the objective of minimizing the distance between the input and the output of the network. It is the usual practice to have a bottleneck latent-space in between with a dimension smaller than the input.  Due to this bottleneck, auto-encoder retains only essential information in the latent space that is required for reconstruction. In a denoising auto-encoder, noise is added to the input image and the network is expected to reconstruct the denoised version of the image. It is shown in the literature that denoising auto-encoders reduce over-fitting and improve generalizabilty of the network compared to regular auto-encoders. As a result, denoising auto-encoders open up the possibility of having a latent dimension larger than the input image dimension \cite{Vincent:2008:AE}.  

Further, our strategy revolves around densely sampling from the latent space. To facilitate this operation, with the intention of having a bounded support for the latent space, we introduce a {\it tanh} activation in the output layer of the encoder. Therefore, support of the latent space is $(-1,1)^d$, where $d$ is the dimension of the latent space. In our implementation, we add zero mean Gaussian white noise with a variance of 0.2 to input images and train the auto-encoder using mean squared error loss as shown below:

$$ l_{\mbox{MSE}} = ||x - \mbox{De}(\mbox{En}(x+n)) ||_2^2,$$

\noindent where $x$ is an input image and $n \sim \mathcal{N}(0,0.2)$. In addition, adversarial loss terms introduced in the following sections are also used to learn parameters of the auto-encoder. Since the decoder part of the auto-encoder also acts as the generator of images from latent space, we use the words {\it decoder} and {\it generator} interchangeably in the remainder of the text. 

\noindent{\textbf{Latent Discriminator:}} The motivation of our method is to obtain a latent space where each and every instance from the latent space represents an image from the given class. If representations of the given class are only confined to a sub-region of the latent space, this goal is not possible to achieve. Therefore, we explicitly force latent representations of in-class examples to be distributed uniformly across the latent space. We achieve this using a discriminator operating in the latent space that we call \textit{latent discriminator $D_l$ }. The latent discriminator is trained to differentiate between latent representations of real images of the given class and samples drawn from a $\mathbb{U}(-1,1)^d$ distribution. We consider a loss of the form:
\begin{eqnarray}
 l_{\mbox{latent}}  &&= -(\mathbb{E}_{s \sim \mathbb{U}(-1,1) } [ \log D_l(s) ] + \nonumber\\
  && \mathbb{E}_{x \sim p_x}[\log (1-D_l( \mbox{En}(x+n)))]) \nonumber
\end{eqnarray}
\noindent where, $p_x$ is the distribution of in-class examples. We train the latent discriminator along with the auto-encoder network using $ \max_{\mbox{En}} \min_{D_l} l_{\mbox{latent}} $. Since the latent space is a hyper-cube with support $(-1,1)^d$, at equilibrium, the latent projections of examples from the given class are expected to be distributed evenly following a $\mathbb{U}(-1,1)^d$ distribution.

\noindent{\textbf{Visual Discriminator:}} In order for the network not to represent any out-of-class objects, we propose to sample exhaustively from the latent space and ensure corresponding images are not from out-of class. Since there are no negative classes present during training, this condition is difficult to enforce. Instead, we make sure that all images generated from latent samples are from the same image space distribution as the given class. In order to enforce this constraint, we use a second discriminator that we call {\it visual} discriminator ($D_v$). 

Visual discriminator is trained to differentiate between images of the given class and images generated from random latent samples using the decoder $\mbox{De}(s)$, where $s$ is a random latent sample. We refer to
latter images as \textit{fake images} for the remainder of the paper. When the visual discriminator is fooled, fake images chosen at random in general will look similar to examples from the given class. We evaluate adversarial loss $l_{\mbox{visual}}$ as follows.
\begin{eqnarray}
 l_{\mbox{visual}}  &=& -(\mathbb{E}_{s \sim \mathbb{U}(-1,1) } [ \log D_v(\mbox{De}(s)) ] + \nonumber\\
&& \mathbb{E}_{x \sim p_l}[\log (1-D_v( x))]) \nonumber
\end{eqnarray}

We learn visual discriminator together with the auto-encoder network using $ \max_{\mbox{De}} \min_{D_v} l_{\mbox{visual}} $. 

\begin{figure}
    \centering
\includegraphics[width=80pt]{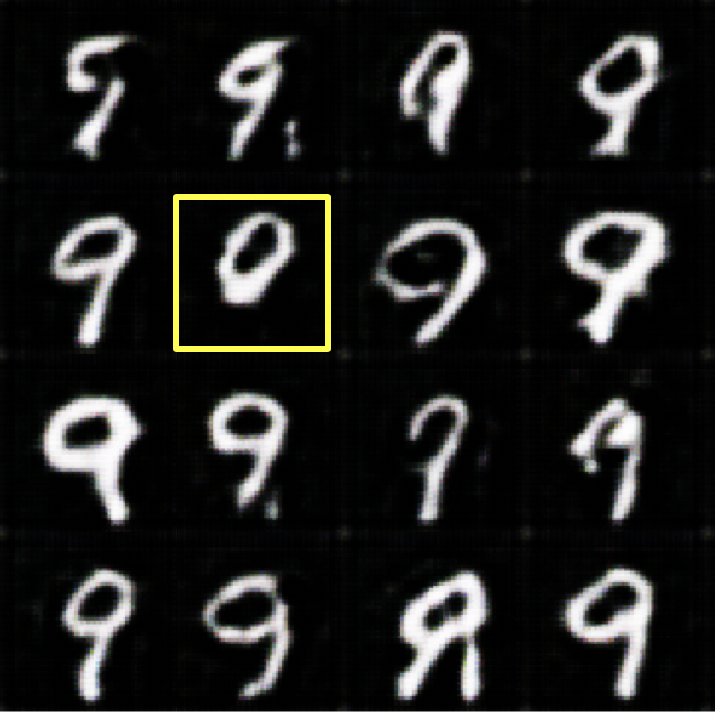}\hskip30pt
\includegraphics[width=80pt]{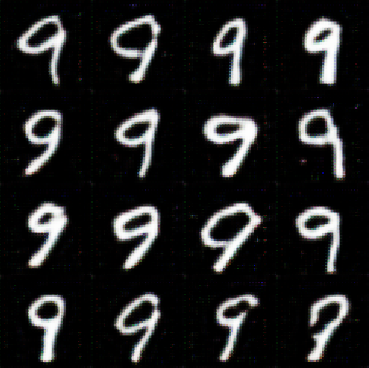}\hskip30pt\\
(a) \hskip 100pt (b)
    \caption{Visualization of generated images from random latent samples when the network is trained (a) without informative-negative mining (b) with informative-negative mining,  for digit 9. In the former case, obtained digits are of a different shape in certain instances. For example, the highlighted generated-image looks like a 0. In the latter case, all generated digits consistently look like a 9. }
    \label{fig:fakenines}
\end{figure}

\noindent \textbf{Informative-negative Mining:} The components described thus far account for the core of the proposed network. Shown in Figure~\ref{fig:fakenines}(a) is a visualization of fake images obtained by jointly training these three sub-networks using digit 9. Figure~\ref{fig:fakenines}(a) suggests that the proposed network is able to generate plausible images of the given class for majority of the random latent samples. However, as indicated in the figure there are few cases where the produced output looks different from the given class. For example, the highlighted digit in Figure~\ref{fig:fakenines}(a)  looks more like a zero than a nine.

This result suggests that despite the proposed training procedure, there are latent space regions that do not produce images of the given class. This is because sampling from all regions in the latent space is impossible during training -- particularly when the latent dimension is large. A naive solution to this problem is to reduce the dimensionality of the latent space. However, with a lower dimension, the amount of detail the network preserves goes down. As a result, although all latent samples produce an in-class image, a very low dimensionality would diminish performance in novelty detection.% Therefor, we abstain from following the naive solution of reducing the dimensionality. 
%Shown in Figure~\ref{nine} is the visualization of the entire latent space trained on digit 9 with a latent dimension of two. All samples produce images belonging to the same class. However, with a lower dimension, the amount of detail network preserves reduces. As a result, all though all latent samples produce an in-class image, a very low dimensionality would diminish performance in novelty detection.% Therefor, we abstain from following the naive solution of reducing the dimensionality. 

%\begin{figure}
%    \centering
%\includegraphics[width=0.7\linewidth]{nines}\hskip30pt
%    \caption{Visualization of the latent space at grid positions when the latent dimension is two for digit 9. All grid points produces plausible digit 9 images. However, mean squared error of known class samples will be low since reconstructed images lack details. }
%    \label{fig:nines}
%\end{figure}

As an alternative, we propose to actively seek regions in the latent space that produce images of poor quality. For the remainder of the paper we refer to these images  as \textit{informative-negative} samples. We use informative-negative samples to train the generator so that it learns to produce good quality in-class images even for these latent samples. However, we continue to use samples chosen at random to train two discriminators, as feeding weaker samples would hinder training of discriminators. In order to find informative-negative samples, first we start with random latent-space samples and use a classifier to assess the quality of the image generated from the sample. The loss of the classifier is used to back-propagate and compute gradients in the latent space. We then take a small step in the direction of the gradient to move to a new point in the latent space where the classifier is confident that the generated image is out-of-class. 

\noindent\textbf{Classifier:} The role of the classifier is to determine how well the given image resembles content of the given class. Ideally such a classifier can be trained using positive and negative examples of a given class. However, since there are no negative training samples available, we train a weaker classifier instead. In the proposed mechanism, if the content belongs to the given class, the classifier deems it positive, and if the content bears no resemblances to the positive class, the classifier deems it negative.

We train the classifier using reconstructions of in-class samples as positives and fake images, those that are generated from random samples in the latent space, as negatives. This classifier is trained independent of other network elements using binary cross entropy loss $l_{\mbox{classifier}}$. In other words, the classifier loss is not considered while learning generator and discriminator parameters. Initially, since the quality of fake samples is poor, the classifier is able to obtain very low loss value. As the quality of fake images improves with training, differentiation becomes harder and it forces the classifier to become smarter.  

It should be noted that the classifier's prediction of a given image as a negative may or may not mean that the given image always corresponds to an informative-negative latent sample. Even if it does not, such images do not hinder the training process at all, and training proceeds as usual.

%\begin{itemize}
%\noindent 1. \textbf{The latent sample is an \textit{informative-negative} one}: This is the intended behaviour of the setup. Back-propagation will move the original sample in the latent space so that it becomes more ambiguous with respect to the classifier.

%\item \textbf{The latent sample is not an \textit{informative-negative} one}: In such instances, resulting images will be similar to other images generated from random samples. Therefore, such images do not hinder the training process at all, and training proceeds as usual.
    
%\end{itemize}

Since the informative-negative classifier does not participate in the GAN game, there is no requirement to balance the capacity of the classifier with the generator (whereas, this is the case for both other discriminators). Therefore, it is possible to make the classifier very strong to increase its confidence in in-class reconstructions. 

\begin{figure}
    \centering
\includegraphics[width=0.7\linewidth]{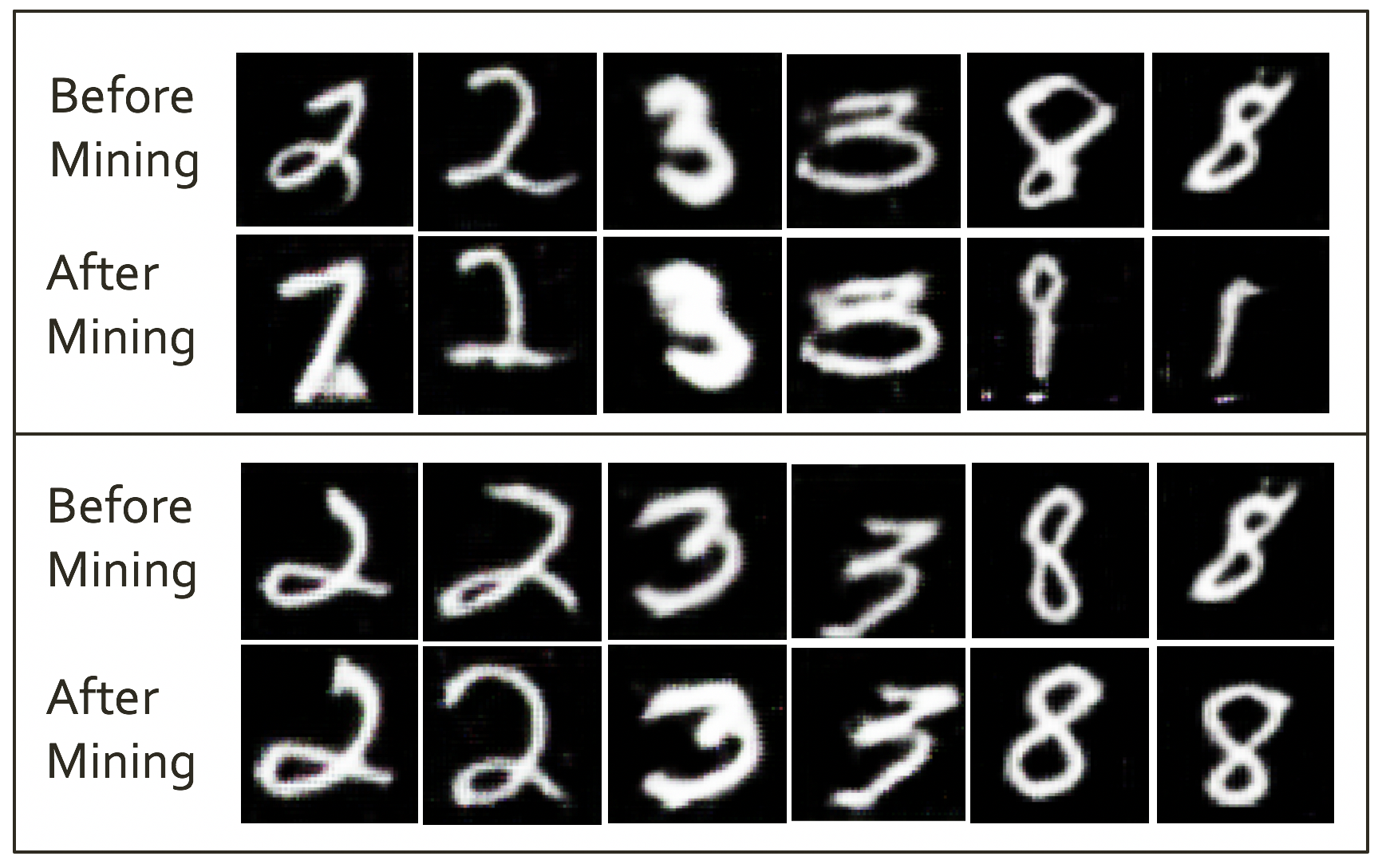}\hskip30pt
    \caption{Informative-negative mining. Shown in the image are image pairs before and after mining process for different digits. In the top row, original images are subjected to substantial changes where they have been converted into a different digits altogether. These are the informative-negatives we are looking for. In the bottom row, the change is not substantial, which means the samples we mined are not informative. However, it still does not hurt our training process.}
    \label{fig:mining}
\end{figure}

Figure~\ref{fig:mining} shows the impact of the informative-negative mining procedure using a few illustrative examples. In the figure, image pairs before and after negative mining are displayed. We have shown cases where the original images are not largely changed in the bottom row. In the top row we have shown a few examples where the input images have been substantially altered as a result of informative-negative mining. For example, the top left sample of digit 2 appears to be a digit 7 after the process. In Figure \ref{fig:fakenines}(b), we show the impact of this procedure by visualizing a few fake images generated from random latent samples for digit 9. It is evident from the figure that informative-negative mining has helped in generating digits of the desired class more consistently across the whole latent space.

\begin{figure*}
    \centering
\includegraphics[width=0.9\linewidth]{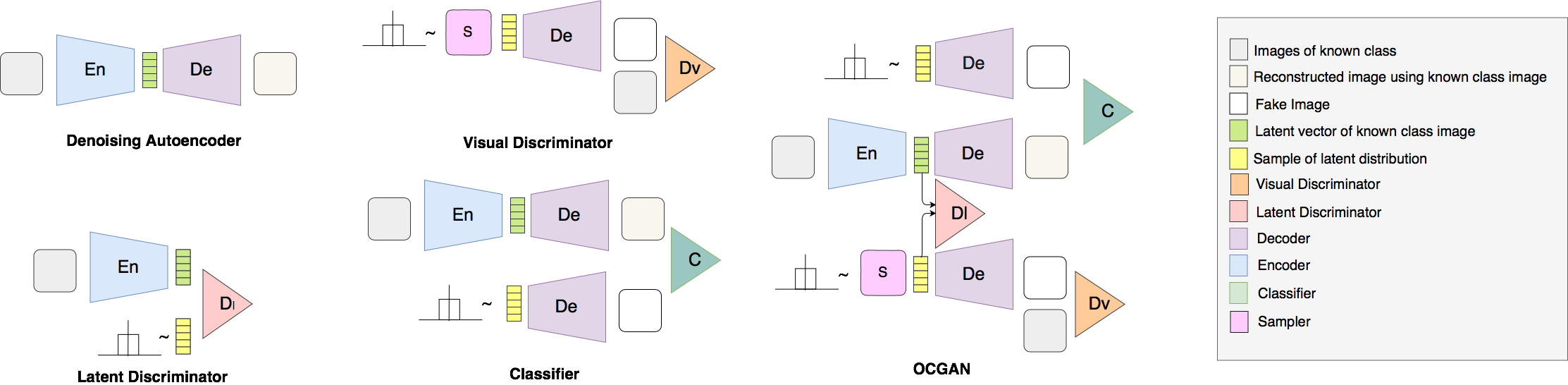}\hskip30pt
    \caption{Illustration of OCGAN architecture: the network consists of four sub-networks : an auto-encoder, two discriminators and a classifier. }
    \label{fig:model}
\end{figure*}

\noindent \textbf{Full OCGAN Model:} The full network of OCGAN and the breakdown of each individual component of the proposed network is shown in Figure~\ref{fig:model}. The network is trained in two iterative steps. In the first step, all other sub-networks except the classifier network are frozen. The classifier network is trained with reconstructed in-class examples and generated fake examples. Then, it is frozen and the auto-encoder and two discriminators are trained adversarially. The latent discriminator is trained based on latent projections of in-class images and random samples drawn from $\mathbb{U}(-1,1)$ distribution. The visual discriminator is trained using fake images generated from random latent samples and real images from the given class. Discriminators are trained by minimizing the loss $l_{\mbox{latent}}+l_{\mbox{visual}}$.

Prior to each generator step, informative-negative samples are sought in the latent space using a batch of random samples drawn from the $\mathbb{U}(-1,1)$ distribution, and using gradient descent steps from the classifier's loss in the latent space. The auto-encoder is trained using informative-negative samples and latent projections of (noise-injected) real examples of the given class using  $10 \times l_{\mbox{MSE}}+l_{\mbox{visual}}+l_{\mbox{latent}}$. A larger weight is given to the $l_{\mbox{MSE}}$ term to obtain good reconstructions. The coefficient was chosen empirically based on the quality of reconstruction. In our implementation, we started mining for informative-negative samples only after the network started producing fake images of reasonable quality. Steps of the training procedure is summarized in Algorithm~\ref{algorithm1}.

\IncMargin{1em}
\begin{algorithm}
	\caption{Training methodology of the OCGAN model: $D_l$, $D_v$ and $C$ represent the outputs of the latent discriminator, visual discriminator and the classifier respectively. {\it En} and {\it De} are the encoder and the decoder/generator respectively. Real label and fake label are denoted by 1 and 0 respectively. }\label{algorithm1}
	\SetKwData{Left}{left}\SetKwData{This}{this}\SetKwData{Up}{up}
	\SetKwFunction{Union}{Union}\SetKwFunction{FindCompress}{FindCompress}
	\SetKwInOut{Input}{Input}\SetKwInOut{Output}{Output}
	\Input{Set of training data $x$, iteration size $N$, parameter $\lambda$}
	\Output{Models: $\text{En}$, De, C, $D_l$ , $D_v$ }
	\BlankLine

		\For{\text{iteration} 1 \KwTo $\rightarrow$  N }{
		    
            \emph{Classifier update: keep $D_l$, $D_v$, En, De fixed}.\\
            
            $n \longleftarrow \mathcal{N}(0,I)$ \\
            $l_1 \longleftarrow \emph{En}(x+n)$ \\
            $l_2 \longleftarrow \mathbb{U}(-1,1)$\\
			$l_{\emph{classifier}} \longleftarrow C(\emph{De}(l_2),0) + C(\emph{De}(l_1),1)$ \\
			$\emph{Back-propagate} l_{\emph{classifier}}  \emph{ to change } C $\\
            	\BlankLine
			$\emph{Discriminator update:}$\\
			$l_{\emph{latent}} \longleftarrow D_l(l_1,0)+D_l(l_2,1)$\\
			$l_{\emph{visual}} \longleftarrow D_v( \emph{De}(l_2),0)+D_v(x,1)$\\
			$\emph{Back-propagate} l_{\emph{latent}}+l_{\emph{visual}} \emph{ and change } D_l, D_v$
				\BlankLine
			
			\emph{Informative-negative mining : Keep all networks fixed.}\\
				%\For{\text{sub-iteration} 1 \KwTo $\rightarrow 5$ }{
				
				\For{\text{sub-iteration} 1 \KwTo $\rightarrow$ $5$  }{
            $l_{\emph{classifier}}$ $\longleftarrow$ C(\emph{De}$(l_2)$, 1)\\
            \emph{Back-propagate } $l_{\emph{classifier}} \emph{ to change }  l_2$\\
            
			}
				\BlankLine
             \emph{Generator update:} \emph{keep} $D_l, D_v$,\emph{C fixed.}\\
			$l_{\emph{latent}} \longleftarrow D_l(l_1, 1)+D_l(l_2, 0)$\\
						$l_{\emph{visual}} \longleftarrow D_v( \emph{De}(l_2),1)+D_v(x,0)$\\
			$l_{\emph{mse}} \longleftarrow ||x-\emph{De}(l_1)||^2$\\
			\emph{Back-propagate} $l_{\emph{latent}}+l_{\emph{visual}}+\lambda l_{\emph{mse}}~\emph{to change En, De}$
		}\end{algorithm}\DecMargin{1em}

\noindent \textbf{Network Architecture and Hyper-parameter Selection:} The auto-encoder is a symmetric network with three 5 x 5 convolutions with stride 2 followed by three transposed convolutions. All convolutions and transposed-convolutions are followed by batch normalization and leaky {\it ReLU} (with slope 0.2) operations. A {\it tanh} activation was placed immediately after the last convolution layer to restrict support of the latent dimension. We used a base channel size of 64 for the auto-encoder and increased number of channels by a factor of 2 with every layer\footnote{Source code in MXNet is made available at https://github.com/PramuPerera/OCGAN.}. 

The visual discriminator and classifier are networks with three 5 x 5 convolutions with stride 2. Base channel size of two networks were chosen to be 12 and 64 respectively. Latent discriminator is a fully connected network with layers of sizes 128, 64, 32 and 16 respectively. Batch normalization and {\it ReLu} activations were used after each layer in all networks.

At the end of training, we selected the model that resulted in minimum MSE on the validation set for evaluation. Model hyper-parameters such as learning rate, latent space size were chosen based on the MSE of validation set. The number of base channels in each network and coefficient of loss terms were decided based on the plot of training loss of each network component.

\section{Experimental Results}
%In this section, we first describe the evaluation methodology used. Then, we introduce datasets used for testing and conclude the section with experimental results.

\subsection{Evaluation Methodology}

We test the effectiveness of the proposed method using four publicly available multi-class object recognition datasets. In order to simulate a one-class setting, each class at a time is considered as the known class, as proposed in \cite{AND}, \cite{GPND} and \cite{dsvdd}. The network is trained using only samples of the known class. During testing, we treat the union of remaining classes as out-of-class samples. Following previous work, we compare the performance of our method using Area Under the Curve (AUC) of Receiver Operating Characteristics (ROC) curve. Here, we note that there exist two protocols in the literature for one-class novelty detection. 

\noindent \textbf{Protocol 1} : Training is carried out using 80$\%$ of in-class samples. The remaining 20$\%$ of in-class data is used for testing. Negative test samples are randomly selected so that they constitute half of the test set.

\noindent \textbf{Protocol 2} : Use the training-testing splits of the given dataset to conduct training. Training split of the known class is used for training / validation. Testing data of all classes are used for testing.

The work of 
\cite{GPND} used the $2^{nd}$ protocol to evaluate their performance in MNIST\cite{MNIST}, FMNIST\cite{fmnist} and COIL100\cite{COIL} datasets, whereas the authors of \cite{AND} and \cite{dsvdd} chose the $1^{st}$ protocol on MNIST and CIFAR10\cite{CIFAR} datasets. We compare our method on these baselines using the relevant protocol for fair comparison.

\begin{figure*}
    \centering
\includegraphics[width=0.9\linewidth]{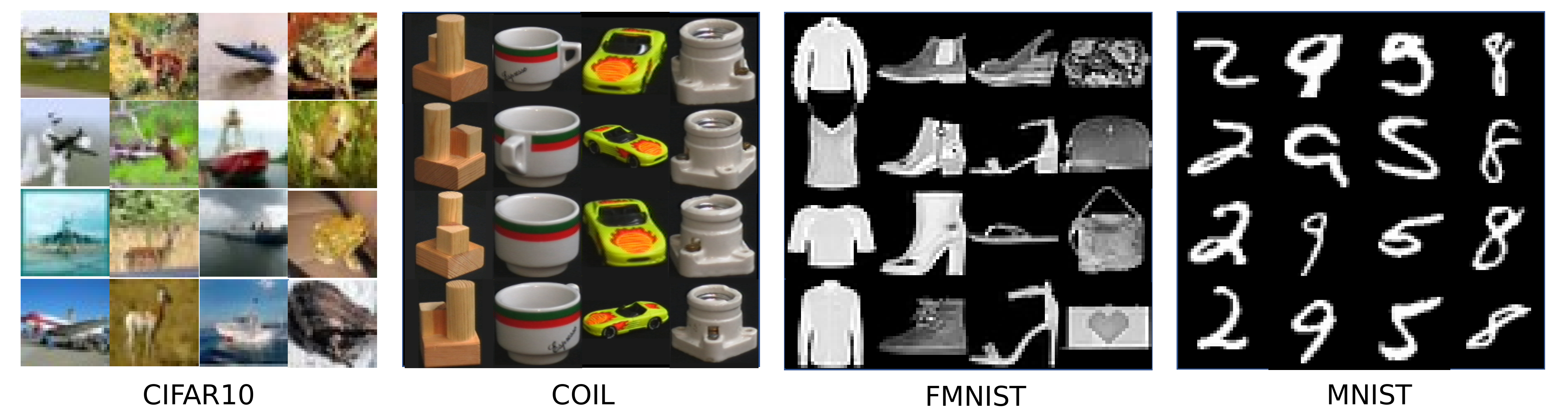}\hskip30pt
    \caption{Representative images from the datasets used for evaluation. Images in each column belong to the same class. }
    \label{fig:datasets}
\end{figure*}

\subsection{Datasets and Experimental Results}
In this section we briefly introduce each dataset used for evaluation and present experimental results for the proposed method. In Figure~\ref{fig:datasets}, a few representative examples from the considered datasets are shown. We tabulate results corresponding to Protocol 1 in Table~\ref{tbl:proto2}
 and results of protocol 2 in Tables \ref{tbl:mnist1} and \ref{tbl:cifar1}. 
 
\noindent  \textbf{COIL100 :} COIL100 is a multi-class dataset where each object class is captured using multiple different poses. There are 100 image classes in the dataset with a few images per class (typically less than hundred). Figure~\ref{fig:datasets} suggests that the intra-class difference is very small for this dataset. As a result, all considered method produces high AUC values for protocol 1 as shown in Table~\ref{tbl:proto2}. Our proposed method of OCGAN records 0.995 AUC, surpassing \cite{GPND} which reported AUC of 0.968.

\noindent  \textbf{fMNIST :} fMNIST is intended to be a replacement for MNIST, where the dataset comprises of 28$\times$28 images of fashion apparels/accessories. As evident from  Figure~\ref{fig:datasets}, fMNIST is a more challenging dataset compared to both COIL100 and MNIST, since there is considerable amount of intra-class variances.  The proposed method improves novelty detection performance by over 2\% compared to \cite{GPND} for this dataset, using protocol 1.

\noindent  \textbf{MNIST :} MNIST dataset contains hand-written digits from 0-9 with a $28\times28$ resolution. This dataset has been widely used to benchmark one-class novelty detection results. In terms of complexity, it is an easier dataset compared to fMNIST, but more challenging than COIL100. We report performances of the proposed method on this dataset using both protocols.

When protocol 1 was used, our OCGAN model yielded an improvement of about 3\% compared to state-of-the-art \cite{GPND} method. As shown in Table~\ref{tbl:mnist1}, when protocol 2 is used, our method has not only registered a better average AUC value, it has reported best AUC for individual classes in 9 out of 10 classes. 

\noindent  \textbf{CIFAR10 :} CIFAR10 is an object recognition dataset that consists of images from 10 classes. Out of the considered datasets, CIFAR10 is the most challenging dataset due to it diverse content and complexity. Specifically, it should be noted that all other datasets are very well aligned, without a background. In comparison, CIFAR10 is not an aligned dataset and it contains objects of the given class across very different settings. As a result, one-class novelty detection results for this dataset are comparatively weaker for all methods. Out of the baseline methods, \cite{dsvdd} has done considerably better than other methods. Following their work, we carried out the same pre-processing in our experiments. In addition, we subtracted the class-mean image from all training and testing images. We obtained comparable results to deep-SVDD with the proposed method where we recorded average AUC of 0.6566.

\begin{table}[]
	\centering
	\caption{Mean One-class novelty detection using Protocol 1.}
	\label{tbl:proto2}

\begin{tabular}{|l|l|l|l|}
\hline
 &MNIST            & COIL  & fMNIST       \\ \hline
ALOCC DR \cite{cvpr2018} & 0.88  & 0.809  & 0.753 \\ \hline
ALOCC D \cite{cvpr2018}  & 0.82  & 0.686  & 0.601 \\ \hline
DCAE    \cite{AE}         & 0.899 & 0.949  & 0.908 \\ \hline
GPND \cite{GPND}     & 0.932 & 0.968  & 0.901 \\ \hline
OCGAN             & \textbf{0.977} & \textbf{0.995}  & \textbf{0.924} \\ \hline
\end{tabular}
\end{table}

\begin{table*}[]
	\centering
	\caption{One-class novelty detection results for MNIST dataset using Protocol 2.}
	\label{tbl:mnist1}
	\resizebox{1.0\linewidth}{!}{%Done
	\begin{tabular}{|l|l|l|l|l|l|l|l|l|l|l|l|}
\hline
\hline
        & 0     & 1     & 2     & 3     & 4     & 5     & 6     & 7     & 8     & 9     & MEAN    \\ \hline
OCSVM \cite{Scholkopf:2001:ESH:1119748.1119749}  & 0.988 & \textbf{0.999} & 0.902 & 0.950 & 0.955 & 0.968 & 0.978 & 0.965 & 0.853 & 0.955 & 0.9513 \\ \hline
KDE \cite{Bishop:2006:PRM:1162264}    & 0.885 & 0.996 & 0.710 & 0.693 & 0.844 & 0.776 & 0.861 & 0.884 & 0.669 & 0.825 & 0.8143 \\ \hline
DAE \cite{MNISTAUTO}    & 0.894 & \textbf{0.999} & 0.792 & 0.851 & 0.888 & 0.819 & 0.944 & 0.922 & 0.740 & 0.917 & 0.8766 \\ \hline
VAE \cite{Variational_autoencoders}    & 0.997 & \textbf{0.999} & 0.936 & 0.959 & 0.973 & 0.964 & 0.993 & 0.976 & 0.923 & 0.976 & 0.9696 \\ \hline
Pix CNN \cite{NIPS2016_6527} & 0.531 & 0.995 & 0.476 & 0.517 & 0.739 & 0.542 & 0.592 & 0.789 & 0.340 & 0.662 & 0.6183 \\ \hline
GAN \cite{IPMI}     & 0.926 & 0.995 & 0.805 & 0.818 & 0.823 & 0.803 & 0.890 & 0.898 & 0.817 & 0.887 & 0.8662 \\ \hline
AND \cite{AND}     & 0.984 & 0.995 & \textbf{0.947} & 0.952 & 0.960 & 0.971 & \textbf{0.991} & 0.970 & 0.922 & 0.979 & 0.9671 \\ \hline
AnoGAN \cite{IPMI} & 0.966 & 0.992 & 0.850 & 0.887 & 0.894 & 0.883 & 0.947 & 0.935 & 0.849 & 0.924 & 0.9127 \\ \hline
DSVDD  \cite{dsvdd}  & 0.980 & 0.997 & 0.917 & 0.919 & 0.949 & 0.885 & 0.983 & 0.946 & \textbf{0.939} & 0.965 & 0.9480 \\ \hline
OCGAN    & \textbf{0.998} & \textbf{0.999} & 0.942 & \textbf{0.963} & \textbf{0.975} & \textbf{0.980} & \textbf{0.991} & \textbf{0.981} & \textbf{0.939} & \textbf{0.981} & \textbf{0.9750} \\ \hline
	
				\end{tabular}}
\end{table*}

\begin{table*}[!h]
	\centering
	\caption{One-class novelty detection results for CIFAR10 dataset using Protocol 2. Plane and Car classes are annotated as Airplane and Automobile in CIFAR10.}
	\label{tbl:cifar1}
	\resizebox{1.0\linewidth}{!}{%Done
		\begin{tabular}{|c|c|c|c|c|c|c|c|c|c|c|c|}
			\hline
			& PLANE  & CAR  & BIRD  & CAT   & DEER  & DOG   & FROG  & HORSE & SHIP  &  TRUCK     &   MEAN            \\ \hline
			\hline
OCSVM \cite{Scholkopf:2001:ESH:1119748.1119749}    & 0.630      & 0.440 & 0.649 & 0.487 & 0.735 & 0.500 & 0.725 & 0.533 & 0.649 & 0.508 & 0.5856 \\ \hline
KDE \cite{Bishop:2006:PRM:1162264}     & 0.658      & 0.520 & 0.657 & 0.497 & 0.727 & 0.496 & \textbf{0.758} & 0.564 & 0.680 & 0.540 & 0.6097 \\ \hline
DAE \cite{MNISTAUTO}     & 0.411      & 0.478 & 0.616 & 0.562 & 0.728 & 0.513 & 0.688 & 0.497 & 0.487 & 0.378 & 0.5358 \\ \hline
VAE \cite{Variational_autoencoders}     & 0.700      & 0.386 & \textbf{0.679} & 0.535 & \textbf{0.748} & 0.523 & 0.687 & 0.493 & 0.696 & 0.386 & 0.5833 \\ \hline
Pix CNN \cite{NIPS2016_6527}  & \textbf{0.788}      & 0.428 & 0.617 & 0.574 & 0.511 & 0.571 & 0.422 & 0.454 & 0.715 & 0.426 & 0.5506 \\ \hline
GAN \cite{IPMI}     & 0.708      & 0.458 & 0.664 & 0.510 & 0.722 & 0.505 & 0.707 & 0.471 & 0.713 & 0.458 & 0.5916 \\ \hline
AND \cite{AND}     & 0.717      & 0.494 & 0.662 & 0.527 & 0.736 & 0.504 & 0.726 & 0.560 & 0.680 & 0.566 & 0.6172 \\ \hline
AnoGAN \cite{IPMI}  & 0.671      & 0.547 & 0.529 & 0.545 & 0.651 & 0.603 & 0.585 & 0.625 & 0.758 & 0.665 & 0.6179 \\ \hline
DSVDD \cite{dsvdd}   & 0.617      & \textbf{0.659} & 0.508 & 0.591 & 0.609 & \textbf{0.657} & 0.677 & \textbf{0.673} & 0.759 & \textbf{0.731} & 0.6481 \\ \hline
OCGAN     & 0.757      & 0.531 & 0.640 & \textbf{0.620} & 0.723 & 0.620 & 0.723 & 0.575 & \textbf{0.820} & 0.554 & \textbf{0.6566} \\ \hline
			
				\end{tabular}}
\end{table*}

\subsection{Ablation Study}
In order to investigate the effectiveness of each additional component of the proposed work, we carried an ablation study using the MNIST dataset. Specifically, we consider four scenarios. In the first scenario we consider only the auto-encoder. In the second and third scenarios, we use auto-encoder with the visual and latent discriminators respectively. In the final scenario, we consider the full proposed model, OCGAN. Mean AUC for each class of MNIST dataset is tabulated in Table~\ref{tbl:ablation}. 

We note that the AUC value obtained for the auto-encoder is already high at 0.957. Therefore even slightest of improvement from this point is significant. When a latent discriminator is
introduced, performance of the system improves marginally by 0.2\%. When a visual discriminator is added on top, the performance improves further by 1\%.  When informative-negative mining as added, performance is further improved by a 0.4\%.

\begin{table}[!h]
	\centering
	\caption{Ablation study for OCGAN performed on MNIST. }
	\label{tbl:ablation}
%Done
		\begin{tabular}{|l|l|}
			\hline
		Without any Discriminators      & 0.957 \\ \hline
With latent Discriminator       & 0.959 \\ \hline
With two Discriminators         & 0.971 \\ \hline
Two Discriminators + Classifier & 0.975 \\ \hline
			
				\end{tabular}
\end{table}

\section{Conclusion}
In this work we dived deep into mechanics of reconstruction-error based novelty detection. We showed that a network trained on a single class is capable of representing some out-of-class examples, given that in-class objects are sufficiently diverse. In order to combat this issue we introduce a latent-space-sampling-based network-learning procedure. First we restricted the latent space to be bounded and forced latent projections of in-class population to be distributed evenly in the latent space using a latent discriminator. Then, we sampled from the latent space and ensured using a visual discriminator that any random latent sample generates an image from the same class. Finally, in an attempt to reduce false positives we introduced an informative-negative mining procedure. We showed that our OCGAN model outperforms many recently proposed one-class novelty detection methods on four publicly available datasets. Further, by performing an ablation study we showed that each component of the proposed method is important for the functionality of the system. 

Experimental results suggest that the proposed method is effective especially when a single concept is present in images as is the case with COIL, MNIST and fMNIST datasets. In future work  we aim to generalize OCGANs to natural image datasets with more complex structure. Further, we wish to investigate their applicability to video novelty detection.

{\small
\bibliographystyle{ieee}
\bibliography{egbib}

\begin{thebibliography}{10}\itemsep=-1pt

\bibitem{AND}
D. {Abati}, A. {Porrello}, S. {Calderara}, and R. {Cucchiara}.
\newblock {AND: Autoregressive Novelty Detectors}.
\newblock In {\em 2019 {IEEE} Conference on Computer Vision and Pattern
  Recognition}, 2019.

\bibitem{Bishop:2006:PRM:1162264}
Christopher~M. Bishop.
\newblock {\em Pattern Recognition and Machine Learning (Information Science
  and Statistics)}.
\newblock 2006.

\bibitem{goodfellow2014generative}
Ian Goodfellow, Jean Pouget-Abadie, Mehdi Mirza, Bing Xu, David Warde-Farley,
  Sherjil Ozair, Aaron Courville, and Yoshua Bengio.
\newblock Generative adversarial nets.
\newblock In {\em Advances in neural information processing systems}, pages
  2672--2680, 2014.

\bibitem{MNISTAUTO}
Raia Hadsell, Sumit Chopra, and Yann Lecun.
\newblock Dimensionality reduction by learning an invariant mapping.
\newblock In {\em In Proc. Computer Vision and Pattern Recognition Conference
  (CVPR’06}. IEEE Press, 2006.

\bibitem{HOFFMANN2007863}
Heiko Hoffmann.
\newblock Kernel pca for novelty detection.
\newblock {\em Pattern Recognition}, 40(3):863 -- 874, 2007.

\bibitem{Variational_autoencoders}
Diederik~P Kingma and Max Welling.
\newblock Auto-encoding variational bayes.
\newblock In {\em International Conference on Learning Representations}.

\bibitem{CIFAR}
Alex Krizhevsky, Vinod Nair, and Geoffrey Hinton.
\newblock Cifar.

\bibitem{MNIST}
Yann LeCun and Corinna Cortes.
\newblock {MNIST} handwritten digit database.

\bibitem{mirza2014conditional}
Mehdi Mirza and Simon Osindero.
\newblock Conditional generative adversarial nets.
\newblock {\em arXiv preprint arXiv:1411.1784}, 2014.

\bibitem{Napierala2016}
Krystyna Napierala and Jerzy Stefanowski.
\newblock Types of minority class examples and their influence on learning
  classifiers from imbalanced data.
\newblock {\em Journal of Intelligent Information Systems}, 46(3):563--597, Jun
  2016.

\bibitem{COIL}
Sameer~A. Nene, Shree~K. Nayar, and Hiroshi Murase.
\newblock Columbia object image library (coil-20.
\newblock Technical report, 1996.

\bibitem{oza2019active}
Poojan Oza and Vishal~M Patel.
\newblock Active authentication using an autoencoder regularized cnn-based
  one-class classifier.
\newblock In {\em 2019 14th IEEE International Conference on Automatic Face \&
  Gesture Recognition (FG 2019)}. IEEE, 2019.

\bibitem{oza2019one}
Poojan Oza and Vishal~M Patel.
\newblock One-class convolutional neural network.
\newblock {\em IEEE Signal Processing Letters}, 26(2):277--281, 2019.

\bibitem{btasoc}
Pramuditha Perera and Vishal~M Patel.
\newblock Dual-minimax probability machines for one-class mobile active
  authentication.
\newblock In {\em IEEE Conference on Biometrics: Theory, Applications, and
  Systems (BTAS)}. IEEE, 2018.

\bibitem{ocfeatures}
P. {Perera} and V.~M. {Patel}.
\newblock {Learning Deep Features for One-Class Classification}.
\newblock {\em ArXiv e-prints}, Jan. 2018.

\bibitem{GPND}
S. {Pidhorskyi}, R. {Almohsen}, D.~A {Adjeroh}, and G. {Doretto}.
\newblock {Generative Probabilistic Novelty Detection with Adversarial
  Autoencoders}.
\newblock In {\em Advances in Neural Information Processing Systems}, 2018.

\bibitem{DCGAN}
Alec Radford, Luke Metz, and Soumith Chintala.
\newblock Unsupervised representation learning with deep convolutional
  generative adversarial networks.
\newblock {\em CoRR}, abs/1511.06434, 2015.

\bibitem{roberts}
Stephen~J Roberts.
\newblock Novelty detection using extreme value statistics.
\newblock {\em IEE Proceedings-Vision, Image and Signal Processing},
  146(3):124--129, 1999.

\bibitem{dsvdd}
Lukas Ruff, Robert Vandermeulen, Nico Goernitz, Lucas Deecke, Shoaib~Ahmed
  Siddiqui, Alexander Binder, Emmanuel M{\"u}ller, and Marius Kloft.
\newblock Deep one-class classification.
\newblock In {\em Proceedings of the 35th International Conference on Machine
  Learning}, pages 4393--4402, 2018.

\bibitem{cvpr2018}
Mohammad Sabokrou, Mohammad Khalooei, Mahmood Fathy, and Ehsan Adeli.
\newblock Adversarially learned one-class classifier for novelty detection.
\newblock In {\em Proceedings of the IEEE Conference on Computer Vision and
  Pattern Recognition}, pages 3379--3388, 2018.

\bibitem{AE}
Mayu Sakurada and Takehisa Yairi.
\newblock Anomaly detection using autoencoders with nonlinear dimensionality
  reduction.
\newblock In {\em Proceedings of the MLSDA 2014 2Nd Workshop on Machine
  Learning for Sensory Data Analysis}, 2014.

\bibitem{Saleh:2013:OAD:2514950.2516141}
Babak Saleh, Ali Farhadi, and Ahmed Elgammal.
\newblock Object-centric anomaly detection by attribute-based reasoning.
\newblock In {\em Proceedings of the 2013 IEEE Conference on Computer Vision
  and Pattern Recognition}, pages 787--794, 2013.

\bibitem{IPMI}
Thomas Schlegl, Philipp Seeb{\"o}ck, Sebastian~M. Waldstein, Ursula
  Schmidt-Erfurth, and Georg Langs.
\newblock Unsupervised anomaly detection with generative adversarial networks
  to guide marker discovery.
\newblock In {\em IPMI}, 2017.

\bibitem{Scholkopf:2001:ESH:1119748.1119749}
Bernhard Sch\"{o}lkopf, John~C. Platt, John~C. Shawe-Taylor, Alex~J. Smola, and
  Robert~C. Williamson.
\newblock Estimating the support of a high-dimensional distribution.
\newblock {\em Neural Comput.}, 13(7):1443--1471, 2001.

\bibitem{Tax:2004:SVD:960091.960109}
David M.~J. Tax and Robert P.~W. Duin.
\newblock Support vector data description.
\newblock {\em Mach. Learn.}, 54(1):45--66, 2004.

\bibitem{NIPS2016_6527}
Aaron van~den Oord, Nal Kalchbrenner, Lasse Espeholt, koray kavukcuoglu, Oriol
  Vinyals, and Alex Graves.
\newblock Conditional image generation with pixelcnn decoders.
\newblock In D.~D. Lee, M. Sugiyama, U.~V. Luxburg, I. Guyon, and R. Garnett,
  editors, {\em Advances in Neural Information Processing Systems 29}, pages
  4790--4798. 2016.

\bibitem{Vincent:2008:AE}
Pascal Vincent, Hugo Larochelle, Yoshua Bengio, and Pierre-Antoine Manzagol.
\newblock Extracting and composing robust features with denoising autoencoders.
\newblock In {\em Proceedings of the 25th International Conference on Machine
  Learning}, pages 1096--1103, 2008.

\bibitem{ICCVoutlier}
Yan Xia, Xudong Cao, Fang Wen, Gang Hua, and Jian Sun.
\newblock Learning discriminative reconstructions for unsupervised outlier
  removal.
\newblock In {\em 2015 {IEEE} International Conference on Computer Vision,
  {ICCV} 2015, Santiago, Chile, December 7-13, 2015}, pages 1511--1519, 2015.

\bibitem{fmnist}
Han Xiao, Kashif Rasul, and Roland Vollgraf.
\newblock Fashion-mnist: a novel image dataset for benchmarking machine
  learning algorithms.
\newblock 2017.

\end{thebibliography}
	}\underline{}
\end{document}